\documentclass[letterpaper]{article} 
\usepackage{aaai2027}  
\nocopyright
\usepackage[hyphens]{url}  
\usepackage{graphicx} 
\usepackage{natbib}  
\usepackage{caption} 
\usepackage{algorithm}
\usepackage{algorithmic}

\usepackage{newfloat}
\usepackage{listings}

\usepackage{amssymb}
\usepackage{amsmath}
\usepackage{colortbl}
\usepackage{xcolor}         

\DeclareCaptionStyle{ruled}{labelfont=normalfont,labelsep=colon,strut=off} 
\floatstyle{ruled}
\newfloat{listing}{tb}{lst}{}
\floatname{listing}{Listing}

\usepackage{booktabs}
\usepackage{multirow,makecell}

\title{Unlearning Is Not Just Erasing: Temporal Decoupling via Generation Inequality}
\author {
    Xunlei Chen\textsuperscript{\rm 1}
    Qirui Ye\textsuperscript{\rm 1},
    Yuang Li\textsuperscript{\rm 1},
    Yi Gong\textsuperscript{\rm 1},\\
    Zhaokun Wang\textsuperscript{\rm 1},
    Wenyi Li\textsuperscript{\rm 1},
    Shiyao Guo\textsuperscript{\rm 1},
    Jinyu Guo\textsuperscript{\rm 2}\
}

\affiliations {
    \textsuperscript{\rm 1}School of Information and Software Engineering, University of Electronic Science and Technology of China\\
    \textsuperscript{\rm 2}School of Computer Science and Engineering, University of Electronic Science and Technology of China \\
    xunlei@std.uestc.edu.cn
}

\begin{document}

\maketitle

\begin{abstract}
Large language models (LLMs) require effective unlearning to address privacy regulations and safety concerns. However, achieving precise forgetting without compromising general utility remains challenging. Existing sequence- and token-level methods penalize target outputs without modeling their context-dependent retrieval paths, which can disrupt linguistic structure or suppress benign knowledge. We present \textbf{ADU}, a fine-grained, training-based framework that shifts unlearning from token erasure to contextual attention-pathway decoupling. Exploiting the functional distinction between local and global attention heads, ADU identifies preplan positions that retrieve persistent sensitive anchors and fixes their candidate paths under the original model. It then trains attention-projection adapters to suppress attention mass along these paths while preserving local-attention structure and retain-set language modeling. Post-training activation exchange tests whether the modified attention-output module transmits the learned forgetting effect. ADU achieves the strongest aggregate performance among evaluated baselines on the TOFU and WMDP benchmarks, including a Forget Quality of \(0.93\) on TOFU. It preserves 87--98\% of model utility (92.9\% on average versus 81.9\% for baselines) while reducing side effects in benign contexts.
\end{abstract}


\section{Introduction}

Large language models (LLMs) have achieved advances in language understanding and generation \citep{Lee2020BioBERT, ranjan2026razor} through model scaling \citep{hu2025exact} and diverse pretraining data \citep{zhao2025qwen3guard}. However, models can reproduce sensitive content \citep{Gong_2026_CVPR}, private information \citep{maini2024tofu}, or unsafe material \citep{Shi2024Safety,shi2025muse}. As the \textit{General Data Protection Regulation} (GDPR) and the \textit{``right to be forgotten''} receive attention \citep{Grynbaum2023Times}, machine unlearning has emerged as a privacy mechanism \citep{zhao2025unlearning}. It aims to reduce the accessibility of requested knowledge while preserving unrelated model capabilities \citep{yu2025unierase}.

Balancing unlearning efficacy and model utility remains fundamentally challenging \citep{pu2026decoding, cha2024learning}. Prompt-based \citep{bhaila2025soft,wang2026cap} and auxiliary-model approaches \citep{geng2025comprehensive, sun2025generative} can change outputs without updating the target model, yet knowledge may remain recoverable through extraction attacks \citep{shah2025unlearning}. Parameter-level unlearning remains necessary when the objective is to modify the deployed model itself, particularly for private or copyrighted content.

\begin{figure}[t]
\centering
\includegraphics[width=\linewidth]{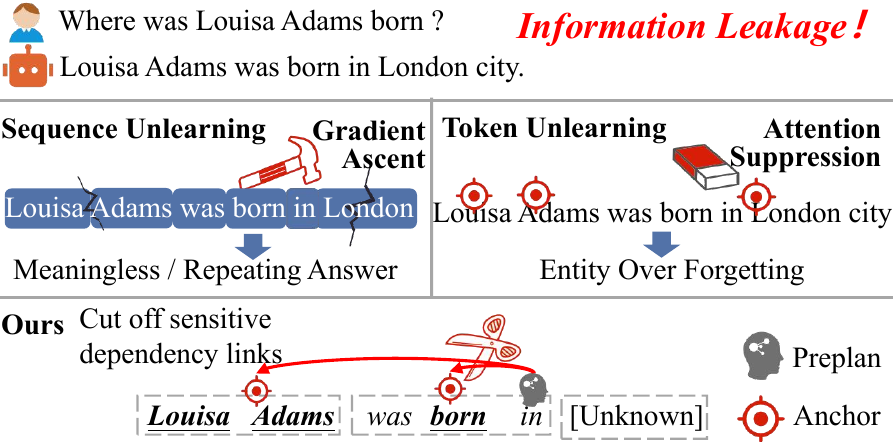}
\caption{Sequence methods minimize the joint probability of the entire QA pair. Token methods blindly suppress the probabilities of individual tokens.
We sever attention pathways that point to the anchor tokens.}
\label{fig:intro}
\end{figure}

\begin{figure*}[t]
\centering
\includegraphics[width=1.0\linewidth]{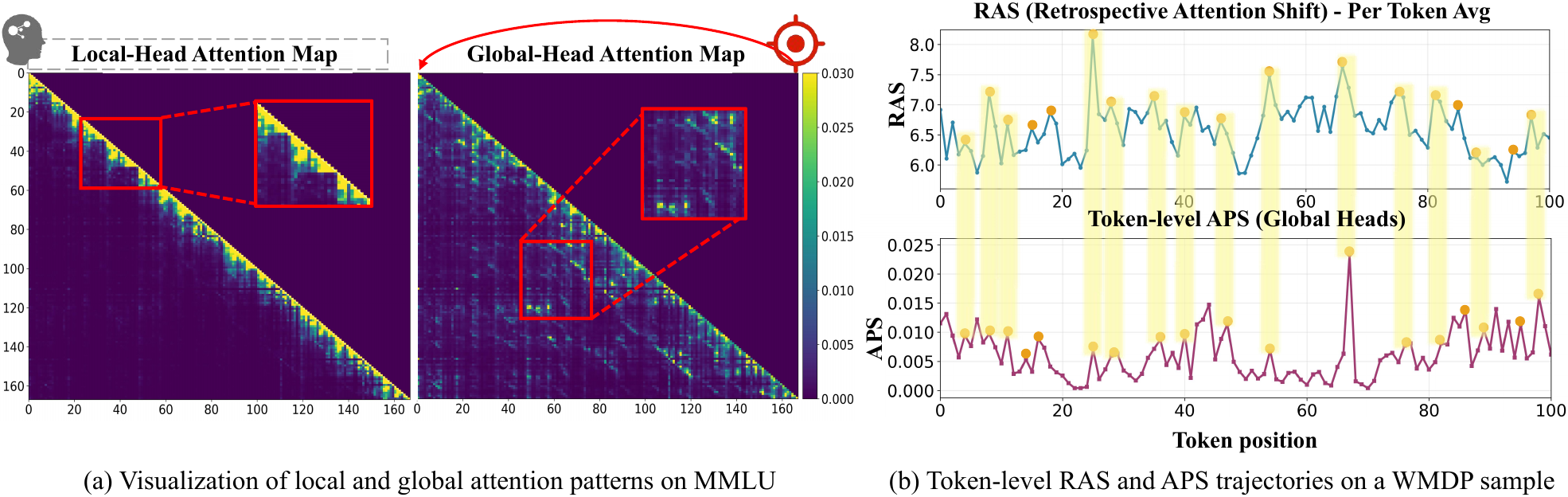}
\caption{Generation inequality and temporal retrieval rhythm. (a) Local heads preserve near-diagonal dependencies, global heads form
long-range patterns. (b) RAS peaks indicate preplan transitions, APS peaks identify persistent anchors, shaded bands mark selected regions. Post-training edge-contribution replacement tests path-specific causal mediation after unlearning.}
\label{fig_temporal_rhythm}
\end{figure*}

Existing parameter-level methods are categorized by their optimization target \citep{zhuang2025seuf,kim2026unlearning}. Sequence-level unlearning treats an entire question--answer sequence as the forget target \citep{liu2025disentangling}. Because its loss spans syntactic and content tokens, it can damage linguistic structure and generation fluency \citep{nguyen2025survey}. Token-level unlearning instead penalizes selected sensitive tokens \citep{jiang2025backdoor} and limits damage by avoiding noncritical positions \citep{yu2025unierase,lee2026direct}. Nevertheless, both approaches define forgetting over visible output targets rather than the internal, context-dependent computation through which sensitive knowledge is retrieved.

This distinction matters because the same entity can be sensitive in one context and benign in another. Static sequence or token penalties may therefore suppress legitimate occurrences, causing excessive forgetting and degraded factual behavior in non-sensitive contexts \citep{tan2025wisdom}. As illustrated in Fig.~\ref{fig:intro}, the desired intervention should target the corresponding retrieval computation rather than erase the entity itself. This raises a natural question: \textit{Can an LLM forget a context-specific retrieval route without destabilizing ordinary generation?}

Motivated by evidence that attention heads contribute unevenly to model behavior~\citep{lin2025critical}, we call their asymmetric temporal roles \emph{generation inequality}. Local heads maintain short-range dependencies, whereas global heads route information from distant, persistently attended tokens. Around semantic transitions, local attention shifts retrospectively at preplan positions, after which global heads retrieve earlier anchors that shape subsequent generation. Average Backward Distance separates the two head groups, while Retrospective Attention Shift and Anchor Persistence Score locate candidate preplans and anchors (Fig.~\ref{fig_temporal_rhythm}).

Based on this rhythm, we propose \textbf{A}ttention \textbf{D}ecoupling \textbf{U}nlearning (\textbf{ADU}), a training-based method for context-specific pathway suppression. ADU computes the head partition and preplan--anchor paths once under the original model and keeps them fixed during training. Attention-projection adapters reduce path mass, while retained language modeling and local-attention preservation constrain collateral changes. The learned decoupling operates in every forward pass; bidirectional edge-contribution replacement runs only after training to test whether the selected preplan--anchor paths causally mediate the resulting forgetting effect.

Unlike representation-level methods such as RMU, token-editing methods such as MET, and broad attention suppression such as ASU, ADU optimizes a context-indexed retrieval pathway rather than a hidden representation or token identity. Attention weights locate candidate routes but are not treated as explanations by themselves: the pathway objective controls transported contributions under bounded values, while bidirectional edge-contribution replacement provides post-training tests of path-specific causal mediation.
Across TOFU, WMDP, and MUSE-Harry Potter, ADU achieves the strongest aggregate forgetting--retention trade-off among the evaluated baselines. It obtains 93\% Forget Quality on TOFU and preserves 87--98\% of model utility, averaging 92.9\% compared with 81.9\% for baselines. These results support contextual pathway suppression as a more targeted alternative to sequence or token erasure. In summary, our contributions are:

1. We formulate LLM unlearning as contextual pathway decoupling and characterize a preplan--anchor temporal pattern arising from the functional specialization of local and global attention heads.

2. We propose ADU, which trains attention-projection adapters to suppress fixed preplan-to-anchor pathways while retaining language modeling and local-attention preservation constrain collateral behavior.

3. We provide conditional theoretical guarantees and extensive evaluations
on WMDP, TOFU, and MUSE-Books, demonstrating state-of-the-art forgetting retention trade-offs and validating the causal role of the targeted pathways.

\begin{figure*}[t]
\centering
\includegraphics[width=1.0\linewidth]{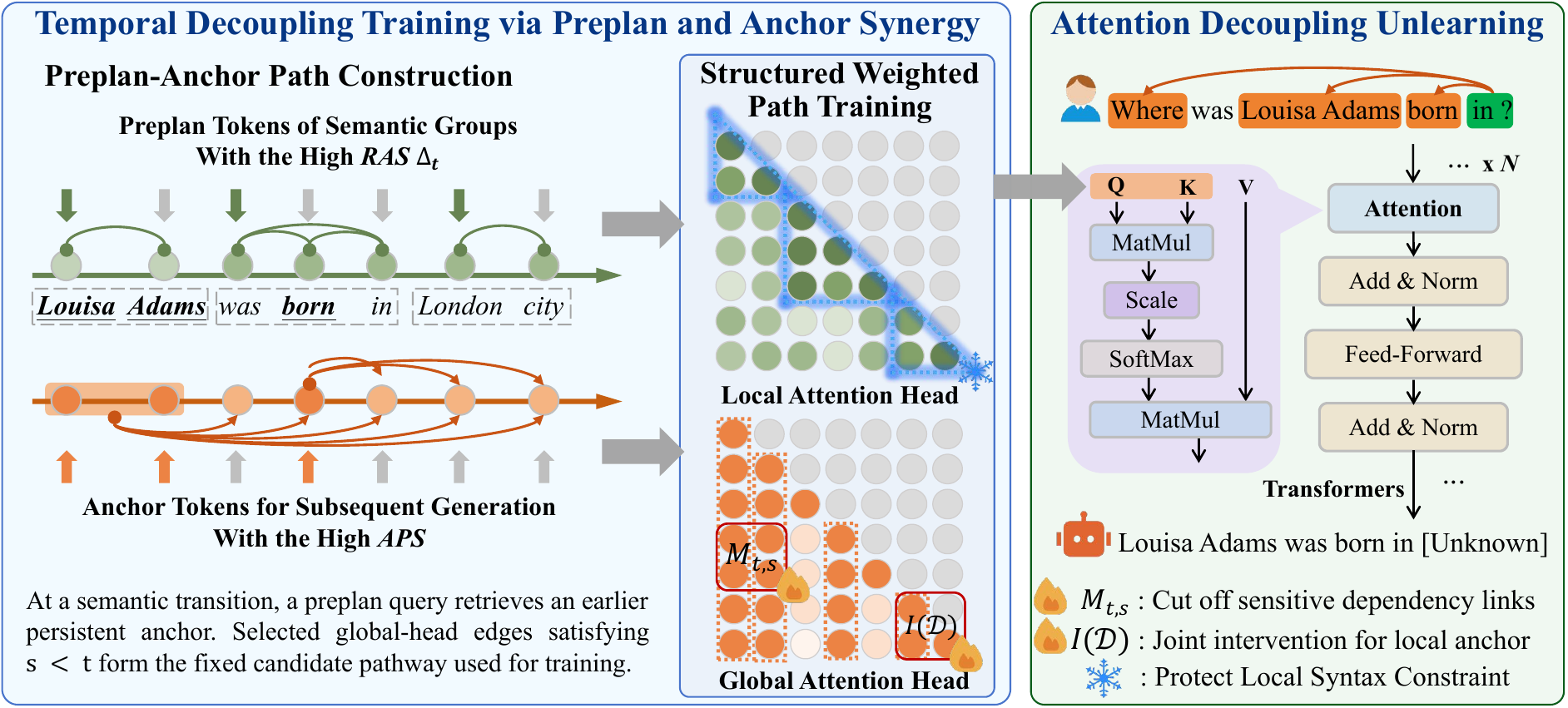}
\caption{Illustration of the workflow. Before predicting the next important token, the preplan token ``in'' queries earlier sensitive anchors through long-range backward attention. After training, ADU suppresses this sensitive dependency path while preserving local anchor structure and maintaining the continuity of local semantic segments.}
\label{fig_adu_framework}
\end{figure*}

\section{Related Work}

\paragraph{Sequence-level Unlearning}
Many parameter-level methods optimize a coarse sequence-level objective. Gradient Ascent (GA) \citep{thudi2022unrolling} directly maximizes the forget-set loss and can produce unstable parameter updates. Negative Preference Optimization (NPO) \citep{zhang2024negative} regularizes this process with a preference objective, but still treats the complete QA pair as one optimization target. Together with their variants \citep{wang2025llm, zhao2024makes, wang2025selective, yang2025cliperase}, such objectives distribute forgetting pressure across many sequence positions, including syntactic function words, without explicitly separating knowledge-bearing content from linguistic structure. Consequently, stronger forgetting can coincide with degraded generation fluency and reduced general capabilities.

\paragraph{Token-level Unlearning}
Token-level methods narrow the target by suppressing specific token logits or redirecting them toward alternatives \citep{yu2025unierase, li2025forget}. Although this avoids some sequence-wide penalties, residual latent semantics may remain, while redirected generation can introduce factual hallucinations. Recently, \citet{tan2025wisdom} identifies important tokens and suppresses attention toward them across the sequence. However, this strategy remains centered on static token salience rather than a context-specific query-to-key retrieval route, limiting contextual discrimination \citep{tran-etal-2025-tokens, jin2025disentangling}. The same entity may therefore be weakened in both sensitive and benign contexts, causing excessive forgetting and disrupting normal knowledge retrieval \citep{yuan2025towards}.

These families differ in granularity but define forgetting mainly over the sequence or token being generated, rather than the contextual computation that retrieves it. ADU instead identifies a recurring preplan--anchor rhythm under the original model and fixes the resulting candidate paths before training. Attention-projection adapters then suppress mass along these paths, while retain language modeling and local-attention preservation constrain collateral changes. Thus, ADU targets context-specific retrieval without directly erasing the sensitive token itself.

\section{Method}
\label{sec_method}

ADU is a training-based unlearning method (see Figure~\ref{fig_adu_framework}). Given an original model \(\theta_0\), a forget set \(D_f\), and a retain set \(D_r\), it learns \(\theta_1=\mathcal U_{\mathrm{ADU}}(\theta_0;D_f,D_r)\) through attention-projection adapters. The learned pathway decoupling acts in every standard forward pass of \(\theta_1\), while counterfactual activation exchange is used only after training to identify the internal computation mediating forgetting.

\subsection{Generation Inequality and Temporal Rhythm}
\label{subsec_temporal_rhythm}

During autoregressive generation, we formalize \emph{generation inequality} as unequal temporal roles: local heads maintain short-range dependencies, whereas global heads retrieve distant, persistently attended tokens. Around semantic transitions, retrospective local shifts mark preplan positions, and persistent global attention identifies earlier anchors that shape generation. This pattern yields the preplan--anchor retrieval hypothesis operationalized below (Figure~\ref{fig_temporal_rhythm}).

Let \(x=(u,o)\) be a context continuation sequence and \(R_x\) the continuation positions. Let \(A_{\theta,t,s}^{(l,h)}(x)\) be the causal attention weight from query position \(t\) to key position \(s\leq t\) in layer \(l\), head \(h\). On a held-out calibration split \(\mathcal D_{\mathrm{cal}}\), the head's average backward distance under the model:
\begin{equation}
d^{(l,h)}
=
\mathbb E_{x\sim\mathcal D_{\mathrm{cal}}}
\left[
\frac{1}{|R_x|}
\sum_{t\in R_x}
\sum_{s\leq t}
A_{\theta_0,t,s}^{(l,h)}(x)(t-s)
\right].
\label{eq_abd}
\end{equation}

The bottom and top \(\rho\) fractions form \(H_{\mathrm{loc}}\) and \(H_{\mathrm{glob}}\), respectively. We use \(\rho=0.3\), compute the partition once from the original model, and keep it fixed during training.

Let \(\bar A_{\theta_0}^{\mathrm{loc}}(x)\) and \(\bar A_{\theta_0}^{\mathrm{glob}}(x)\) denote original-model attention averaged over the fixed local and global head sets. Given a retrospective distance cap \(W\) and a fixed future continuation window \(F_x(s)\), we define
\begin{equation}
\begin{aligned}
r_t
&=
\sum_{s\leq t}
\bar A_{\theta_0,t,s}^{\mathrm{loc}}(x)
\min(t-s,W),\\
\Delta_t
&=
|r_t-r_{t-1}|,
\qquad
a_s
=
\frac{1}{|F_x(s)|}
\sum_{u\in F_x(s)}
\bar A_{\theta_0,u,s}^{\mathrm{glob}}(x).
\end{aligned}
\label{eq_ras_aps}
\end{equation}

Here, \(\Delta_t\) is evaluated at continuation positions with a preceding continuation position, and positions without a valid future window are excluded from APS selection. Large \(\Delta_t\) identifies a candidate transition, whereas large \(a_s\) indicates persistent influence on subsequent queries. For selection ratio \(q\), let \(Q_{1-q}(\Delta;x)\) and \(Q_{1-q}(a;x)\) denote the corresponding within-sequence quantiles. We define
\begin{equation}
\begin{aligned}
T_{\mathrm{pre}}(x)
&=
\left\{
t\in R_x\mid
\Delta_t\geq Q_{1-q}(\Delta;x),\
r_t\geq\tau_{\mathrm{ras}}
\right\},\\
S_{\mathrm{anc}}(x)
&=
\left\{
s\in R_x\mid
a_s\geq Q_{1-q}(a;x)
\right\}
\cap C_f(x).
\end{aligned}
\label{eq_preplan_anchor}
\end{equation}

Here, \(C_f(x)\) contains candidate sensitive positions derived from the forget continuation. Exact window construction and sensitive-position filtering are detailed in Appendix B.
These signals locate candidate retrieval sites; ADU turns the identified rhythm into a trainable mechanism by suppressing the corresponding attention pathway.

\subsection{Pathway Contributions and Causal Mediation}
\label{subsec_causal_formulation}

For each forget sample, \(\mathcal P(x)\) contains tuples
\(e=(l,h,t,s)\) such that \((l,h)\in H_{\mathrm{glob}}\),
\(t\in T_{\mathrm{pre}}(x)\), \(s\in S_{\mathrm{anc}}(x)\), and
\(s<t\), where the last condition enforces causal attention order. We
distinguish the contribution before and after the effective output
projection:
\begin{equation}
\begin{aligned}
\widetilde C_e(\theta,x)
&=
A_{\theta,t,s}^{(l,h)}(x)
V_{\theta,s}^{(l,h)}(x),\\
C_e(\theta,x)
&=
W_{O,\theta}^{(l,h)}
\widetilde C_e(\theta,x).
\end{aligned}
\label{eq_edge_contribution}
\end{equation}

Here, \(\widetilde C_e\) is the head-space contribution manipulated by
the intervention, while \(C_e\) is its residual-stream image controlled
by the pathway objective. For \(a\in\{0,1\}\), define the path-specific
causal mediator as
\begin{equation}
\mathbf M_a(x)
=
\left(
\widetilde C_e(\theta_a,x)
\right)_{e\in\mathcal P(x)}.
\label{eq_causal_mediator}
\end{equation}

Replacing \(\mathbf M_a(x)\) by \(\mathbf M_b(x)\) subtracts the running
model's selected contributions and adds the source model's corresponding
contributions at each affected pre-output-projection head output. The
recipient model retains its own \(W_O\) and every unpatched computation,
and all downstream activations are recomputed. Here, \(a=0\) denotes the
original model and \(a=1\) the trained ADU model.

For theoretical and causal analysis, let \(I_f(x)\) denote the positions
of a sensitive continuation span. For each \(i\in I_f(x)\), let \(y_i\)
be the target token and \(\mathcal C_i(x)\) a prespecified set of
non-target contrasts. Under teacher forcing, the sensitive accessibility
score is
\begin{equation}
Y_\theta(x)
=
\frac{1}{|I_f(x)|}
\sum_{i\in I_f(x)}
s_\theta(y_i,x_{<i}),
\label{eq_sensitive_score}
\end{equation}
where
\[
s_\theta(y_i,x_{<i})
=
z_\theta(y_i\mid x_{<i})
-
\log
\sum_{c\in\mathcal C_i(x)}
\exp z_\theta(c\mid x_{<i}).
\]
This dataset-agnostic score is used only to formalize sensitive
accessibility and counterfactual effects; it is not part of the ADU
training objective.
\begin{equation}
\begin{aligned}
\Delta_f
&=
\mathbb E_{x\sim D_f}
\left[
Y_{\theta_0}(x)-Y_{\theta_1}(x)
\right]
\geq\epsilon_f>0,\\
\Delta_r
&=
\mathbb E_{x\sim D_r}
\left[
\mathcal D_{\mathrm{ret}}
\left(
\pi_{\theta_1}(\cdot\mid x),
\pi_{\theta_0}(\cdot\mid x)
\right)
\right]
\leq\epsilon_r,
\end{aligned}
\label{eq_unlearning_definition}
\end{equation}
where \(\mathcal D_{\mathrm{ret}}\) measures retain-behavior discrepancy.

Let \(Y(a,\mathbf m;x)\) denote the counterfactual score from \(\theta_a\) when its selected head-space contributions are replaced by \(\mathbf m\) before \(W_{O,\theta_a}\). All unpatched computations and parameters stay as \(\theta_a\), and downstream activations are recomputed. Define
\(Y_{ab}(x)=Y(a,\mathbf M_b(x);x)\), where the first index marks the running model and the second the mediator source. Writing
\(\mathbb E_f\) for expectation over \(D_f\), we obtain
\begin{equation}
\begin{aligned}
\mathrm{TE}
&=
\mathbb E_f[Y_{00}-Y_{11}],\\
\mathrm{IE}_{\mathrm{sup}}
&=
\mathbb E_f[Y_{00}-Y_{01}],
&
\mathrm{IE}_{\mathrm{res}}
&=
\mathbb E_f[Y_{10}-Y_{11}].
\end{aligned}
\label{eq_causal_effects}
\end{equation}

Under consistency,
\(Y(a,\mathbf M_a(x);x)=Y_{\theta_a}(x)\), so
\(Y_{aa}=Y_{\theta_a}\) and \(\mathrm{TE}=\Delta_f\). The suppression
effect replaces Base contributions with their ADU counterparts, whereas
the restoration effect replaces ADU contributions with their Base
counterparts. These bidirectional interventions test whether the
selected preplan--anchor contributions causally mediate the learned
forgetting effect. Direct remainders representing additional unpatched
routes are defined in Appendix A.

\subsection{Training Objective and Theoretical Guarantees}
\label{subsec_training_theory}

Let \(N_x=\max(1,|\mathcal P(x)|)\) and \(A_{\theta,e}(x)=A_{\theta,t,s}^{(l,h)}(x)\) for \(e=(l,h,t,s)\). The pathway mass and forget loss are
\begin{equation}
\mathrm{PM}_\theta(x)
=
\frac{1}{N_x}
\sum_{e\in\mathcal P(x)}
A_{\theta,e}(x),
\qquad
\mathcal L_f
=
\mathbb E_{x\sim D_f}
[\mathrm{PM}_\theta(x)].
\label{eq_pathway_loss}
\end{equation}

The pathway indices are computed once under \(\theta_0\) and kept fixed during training; gradients flow through the selected attention values but not through the discrete mask. We combine \(\mathcal L_f\) with language modeling on \(D_r\) and a row-wise loss that preserves original-model local attention on \(D_f\cup D_r\):
\begin{equation}
\mathcal L_{\mathrm{ADU}}
=
\alpha\mathcal L_f
+
(1-\alpha)
\left(
\mathcal L_{\mathrm{lm}}
+
\mathcal L_{\mathrm{loc}}
\right).
\label{eq_adu_objective}
\end{equation}

Here, \(\alpha\in[0,1]\) controls the forget--retain balance. The backbone remains frozen, while LoRA adapters update \(W_Q,W_K,W_V,\) and \(W_O\) in layers containing selected global heads. Samples with \(\mathcal P(x)=\varnothing\) contribute zero to \(\mathcal L_f\). 
The complete retain objective, trainable scope, and empty-path handling
are detailed in Appendix B.

\begin{table*}[t]
\centering
\resizebox{0.78\textwidth}{!}{
\begin{tabular}{l|ccccc}
\toprule
{\textbf{Method}}
& \multicolumn{2}{c}{Forget tasks(\%)}
& \multicolumn{3}{c}{Retain tasks(\%)} \\
\cmidrule(lr){2-3}
\cmidrule(lr){4-6}
\textbf{Llama3.1-8B-Instruct} & \textbf{Bio.$\downarrow$}
& \textbf{Cyber$\downarrow$}
& \textbf{MMLU$\uparrow$}
& \textbf{GSM8K$\uparrow$}
& \textbf{Flu.$\uparrow$} \\
\cmidrule(lr){1-6}
 Base & 71.86 & 45.37 & 68.16 & 67.83 & 3.74 \\
 NPO\_KL\textsuperscript{$\ddagger$}~\citep{zhang2024negative} & 56.38 & 34.32 & 52.37 & 53.55 & 2.79 \\
 RMU\textsuperscript{$\ddagger$}~\citep{li2025wmdp} & 39.55 & 31.75 & 53.43 & 52.64 & 2.90 \\
\cmidrule(lr){1-6}
 ICUL\textsuperscript{$\diamond$}~\citep{pawelczyk2024context} & 41.31 & 30.46 & 60.69 & 58.40 & \textbf{3.58} \\
 ALU\textsuperscript{$\diamond$}~\citep{sanyal2025agents} & 31.87 & \underline{29.94} & \textbf{63.78} & \underline{58.65} & {3.26} \\
\cmidrule(lr){1-6}
 MET\textsuperscript{$\dagger$}~\citep{yu2025unierase} & 34.63 & 31.47 & 52.19 & 54.19 & 3.18 \\
 ASU\textsuperscript{$\dagger$}~\citep{tan2025wisdom} & 34.49 & 33.17 & 58.58 & 57.24 & {3.31} \\
 ALTER\textsuperscript{$\dagger$}~\citep{chen2026alter} & \underline{29.69} & 30.77 & {60.10} & 57.60 & {3.20} \\
 ADU\textsuperscript{$\dagger$} (Ours) & \textbf{27.32} & \textbf{27.97} & \underline{62.84} & \textbf{58.82} & \underline{3.34} \\
\midrule
\textbf{Qwen3-14B}& \textbf{Bio.$\downarrow$}
& \textbf{Cyber$\downarrow$}
& \textbf{MMLU$\uparrow$}
& \textbf{GSM8K$\uparrow$}
& \textbf{Flu.$\uparrow$} \\
\cmidrule(lr){1-6}
Base & 76.07 & 50.98 & 75.18 & 79.25 & 3.80 \\
NPO\_KL\textsuperscript{$\ddagger$}~\citep{zhang2024negative} & 60.59 & 39.85 & 64.88 & 63.50 & 2.88 \\
RMU\textsuperscript{$\ddagger$}~\citep{li2025wmdp} & 43.85 & 36.24 & 66.51 & 64.86 & 3.08 \\
\cmidrule(lr){1-6}
ICUL\textsuperscript{$\diamond$}~\citep{pawelczyk2024context} & 46.82 & 31.50 & 65.22 & 68.37 & \textbf{3.69} \\
ALU\textsuperscript{$\diamond$}~\citep{sanyal2025agents} & \underline{31.71} & \underline{30.38} & 67.67 & 68.81 & \underline{3.58} \\
\cmidrule(lr){1-6}
MET\textsuperscript{$\dagger$}~\citep{yu2025unierase} & 38.28 & 34.53 & 65.49 & 66.29 & 3.27 \\
ASU\textsuperscript{$\dagger$}~\citep{tan2025wisdom} & 32.09 & 33.89 & 69.88 & \underline{71.54} & {3.47} \\
ALTER\textsuperscript{$\dagger$}~\citep{chen2026alter} & 34.24 & 36.58 & \textbf{71.55} & 70.83 & 3.35 \\
ADU\textsuperscript{$\dagger$} (Ours) & \textbf{29.40} & \textbf{29.12} & \underline{70.91} & \textbf{73.28} & {3.57} \\
\bottomrule
\end{tabular}
}
\caption{Multiple-choice accuracy on the forgetting/retention benchmark after unlearning. \textsuperscript{$\dagger$}, \textsuperscript{$\diamond$}, and \textsuperscript{$\ddagger$} denote token-level training, prompt-based methods, and sequence-level training, respectively.}
\label{tab:WMDP_performance_main}
\end{table*}

\paragraph{From pathway training to knowledge suppression.}
If \(\|W_{O,\theta}^{(l,h)}V_{\theta,s}^{(l,h)}(x)\|_2\leq B\) for all selected edges, then
\begin{equation}
\frac{1}{N_x}
\sum_{e\in\mathcal P(x)}
\|C_e(\theta,x)\|_2
\leq
B\,\mathrm{PM}_\theta(x).
\label{eq_surrogate_control}
\end{equation}
Thus, minimizing pathway mass controls the average transported magnitude of selected edge contributions rather than treating attention weights as explanations. These contributions form the path-specific component of the attention-output computation examined by activation exchange.

For an affected query--head row \(j\), let \(p_j\) and \(p'_j\) be its selected-anchor mass before and after pathway decoupling, with \(\delta_j=p_j-p'_j\geq0\). Write
\[
\mathbf o_j(p_j)
=
p_j\mu_{S,j}
+
(1-p_j)\mu_{\bar S,j},
\qquad
\Gamma_j
=
\mu_{S,j}-\mu_{\bar S,j},
\]
where \(\mu_{S,j}\) and \(\mu_{\bar S,j}\) are the normalized selected-anchor and complementary value-output mixtures. A mass-transfer intervention holds these mixtures fixed while reducing \(p_j\), yielding
\(
\mathbf o_j(p'_j)-\mathbf o_j(p_j)
=
-\delta_j\Gamma_j.
\)

Let \(g(\mathbf p)=Y(\mathbf o_1(p_1),\ldots,\mathbf o_J(p_J))\) be the sensitive score induced by the affected attention outputs. Assume that \(g\) is differentiable and, at every point along the intervention path,
\(
\left\langle
\nabla_{\mathbf o_j}g,\Gamma_j
\right\rangle
\geq
\kappa_j>0.
\)
For the retain-discrepancy functional \(g_r\), assume \(L\)-Lipschitz continuity and \(\|\Gamma_j\|_2\leq B_j\). Then
\begin{equation}
\begin{aligned}
g(\mathbf p')-g(\mathbf p)
&\leq
-\sum_j\kappa_j\delta_j,\\
|g_r(\mathbf p')-g_r(\mathbf p)|
&\leq
L\sum_jB_j\delta_j.
\end{aligned}
\label{eq_forget_retain_bounds}
\end{equation}
The first inequality gives a sufficient condition for reducing sensitive log-odds, while the second bounds the retain-discrepancy change attributable to the same intervention. Complete proofs are provided in Appendix~A.

Together, the pathway loss controls attention contributions, directional alignment translates their reduction into lower sensitive log-odds, and the retain objective constrains changes outside the pathway. Bidirectional activation exchange then tests whether the modified attention-output computation mediates the forgetting effect.

\begin{table*}[h]
\centering
\resizebox{0.76\textwidth}{!}{
\begin{tabular}{l|ccc|cc|c}
\toprule
\textbf{Method}
& \multicolumn{3}{c|}{\textbf{TUD}} 
& \multicolumn{2}{c|}{\textbf{NEK}} 
& \textbf{GEK} \\
\cmidrule{2-4} \cmidrule{5-6} \cmidrule{7-7}
\textbf{Llama3.1-8B}
& \textbf{R-L$\downarrow$} 
& \textbf{TR$\uparrow$} 
& \textbf{FQ$\uparrow$} 
& \textbf{R-L$\uparrow$} 
& \textbf{Acc$\uparrow$} 
& \textbf{Acc$\uparrow$} \\
\midrule
NPO\_KL\textsuperscript{$\ddagger$}~\citep{zhang2024negative} & 0.31 & 0.78 & 0.72 & 0.58 & 62.2 & 64.2 \\
RMU\textsuperscript{$\ddagger$}~\citep{li2025wmdp} & 0.19 & 0.91 & {0.88} & 0.62 & 65.1 & 68.3 \\
\midrule
ICUL\textsuperscript{$\diamond$}~\citep{pawelczyk2024context} & 0.17 & 0.94 & 0.54 & 0.61 & 63.8 & 69.0 \\
ALU\textsuperscript{$\diamond$}~\citep{sanyal2025agents} & \underline{0.13} & \underline{0.95} & 0.67 & 0.64 & 66.7 & 70.6 \\
\midrule
MET\textsuperscript{$\dagger$}~\citep{yu2025unierase} & 0.18 & 0.88 & \underline{0.92} & 0.64 & 63.3 & 70.2 \\
ASU\textsuperscript{$\dagger$}~\citep{tan2025wisdom} & 0.16 & 0.93 & 0.87 & \underline{0.66} & \underline{69.6} & 71.0 \\
ALTER\textsuperscript{$\dagger$}~\citep{chen2026alter} & 0.14 & 0.91 & 0.81 & 0.60 & 67.2 & \underline{71.3} \\
ADU\textsuperscript{$\dagger$} (Ours) & \textbf{0.11} & \textbf{0.96} & \textbf{0.93} & \textbf{0.69} & \textbf{70.8} & \textbf{72.3} \\
\bottomrule
\end{tabular}
}
\caption{Performance comparison on TOFU (10\%) with Llama3.1-8B-Instruct.}
\label{tab:tofu_performance}
\end{table*}

\begin{table}[t]
\centering
\begin{tabular}{lcccc}
\toprule
\textbf{Method} & \textbf{BLEU$\downarrow$} & \textbf{R-L$\downarrow$} & \textbf{MMLU$\uparrow$} & \textbf{Flu.$\uparrow$} \\
\midrule

Original  & 74.80 & 85.14 & 46.33 & 3.63 \\
NPO\textsuperscript{$\ddagger$}  & \textbf{1.55} & 14.08 & 42.70 & 2.96 \\
WHP\textsuperscript{$\ddagger$}  & 23.68 & 17.93 & 43.49 & 2.52 \\
ALU\textsuperscript{$\diamond$}  & 7.21 & 14.85 & 44.34 & {3.27} \\
ICUL\textsuperscript{$\diamond$} & 27.50 & 25.89 & 44.02 & \textbf{3.34} \\
ALTER\textsuperscript{$\dagger$} & 6.96 & \underline{10.40} & 43.84 & 2.32 \\
Ours\textsuperscript{$\dagger$}  & \underline{4.78} & \textbf{9.49} & \textbf{45.64} & \underline{3.29} \\

\bottomrule
\end{tabular}

\caption{Results on MUSE-Harry Potter with Llama2-7B.}
\label{tab:HP_performance}
\end{table}
\section{Experiment}
\subsection{Experiment Settings}
\paragraph{Datasets}
We evaluate our method on three benchmarks. WMDP \citep{li2025wmdp} verifies forgetting and retention effectiveness by assessing model knowledge in sensitive domains such as biosafety and cybersecurity. MUSE‑Harry Potter \citep{shi2024muse} assesses copyright unlearning: models are first fine‑tuned on the Harry Potter book content to memorize it, then unlearned to forget that content. TOFU \citep{maini2024tofu} examines boundary preservation between the forget set and its neighboring retain set, simulating synthetic and real-world unlearning scenarios.
We note that all three benchmarks involve extended, context-rich responses where the model must retrieve factual knowledge through multi-token generation--precisely the regime where preplan-to-anchor attention patterns emerge and ADU's pathway decoupling is most effective. 
To test general ability, we apply MMLU \citep{hendrycks2021measuring} for fact answering and GSM8K \citep{cobbe2021training} for math reasoning. Datasets details and configurations are provided in Appendix C.1.

\paragraph{Metrics}
For WMDP, we report multiple choice accuracy on Bio and Cyber as forgetting metrics, where lower values indicate stronger forgetting. MMLU and GSM8K are used as retention metrics, where higher values indicate stronger utility preservation. For MUSE Harry Potter, we report BLEU and ROUGE-L to measure textual overlap with copyrighted content, together with MMLU and fluency for utility. For TOFU, we report ROUGE-L on the target unlearned data, Top 5 exclusion rate, Forget Quality, ROUGE-L on neighboring knowledge, neighboring accuracy, and general knowledge accuracy. Fluency is evaluated by GPT-4o on a 1 to 5 scale. We also report Forgetting Performance (FP) and Retaining Performance (RP) in analysis, where FP is the average of WMDP Bio and Cyber, and RP is the average of MMLU and GSM8K. Details are provided in Appendix C.2.

\paragraph{Baselines}
We compare with sequence-level training, prompt-based methods, and token-level training methods. Sequence-level training includes NPO\_KL~\citep{zhang2024negative}, and Representation Misdirection for Unlearning (RMU)~\citep{li2025wmdp}. Prompt-based methods include ICUL~\citep{pawelczyk2024context} and ALU~\citep{sanyal2025agents}. Token-level training includes Model Edit Token (MET)~\citep{yu2025unierase}, Attention Shift Unlearning (ASU)~\citep{tan2025wisdom}, and Hydra Suppress Unlearning (ALTER)~\citep{chen2026alter}.

\subsection{Main Result}
\paragraph{Forgetting-Retention Effectiveness} 

We report WMDP forgetting and general retention results (Table~\ref {tab:WMDP_performance_main}). On Llama3.1-8B-Instruct, ADU reduces Bio accuracy from 71.86 to 27.32 and Cyber accuracy from 45.37 to 27.97, retaining MMLU and GSM8K at 62.84 and 58.82.
On Qwen3-14B, ADU achieves the lowest Bio and Cyber accuracy and the best GSM8K retention among unlearning methods; prompt-based ALU ranks second on Cyber forgetting without modifying model parameters. These results show that ADU improves trainable forgetting and retention trade off instead of optimizing one forgetting metric at the cost of utility.
Table~\ref {tab:HP_performance} evaluates copyright unlearning on MUSE Harry Potter. ADU achieves the lowest ROUGE-L among unlearning methods and strongest MMLU retention. Although NPO gives lower BLEU, its MMLU drops to 42.70, while ADU keeps MMLU at 45.64 and fluency at 3.29. This pattern supports the pathway view. ADU weakens the route to memorized content while avoiding broad degradation of general next token behavior.
Settings and costs are in Appendix~C.4.

\begin{table}[t]
\centering
\begin{tabular}{lccc}
\toprule
\textbf{Setting} & \textbf{Avg$\downarrow$} & \textbf{MMLU$\uparrow$} & \textbf{GSM8K$\uparrow$} \\
\midrule
ADU                  & 27.65 & 62.84 & 58.82 \\
w/o pathway loss     & 47.79 & 63.22 & 58.44 \\
w/o retain objective & 25.68 & 56.32 & 52.10 \\
random heads         & 34.79 & 59.26 & 55.28 \\
w/o anchor filter    & 26.19 & 58.40 & 54.32 \\
w/o RAS preplan      & 32.38 & 60.05 & 55.99 \\

\bottomrule
\end{tabular}

\caption{Component ablation on Llama3.1-8B-Instruct. Avg denotes the average of WMDP Bio and Cyber.}
\label{tab_component_ablation}
\end{table}

\paragraph{Boundary Preservation} 
To further verify the impact of unlearning methods on neighboring and retained knowledge, we conducted experiments on TOFU (10\%), as shown in Table~\ref{tab:tofu_performance}.
Sequence-level training methods struggle to balance the trade-off between forgetting and model utility.
Prompt-based methods provide stronger retention, but their forgetting and neighboring preservation remain unstable.
Although token-level training methods improve this balance, existing variants may still disrupt semantic dependencies shared with neighboring facts. For example, ASU obtains 69.6\% NEK accuracy, but its TUD R-L remains 0.16.
By severing hazardous retrieval pathways while preserving adjacent semantic pathways, ADU achieves the best TUD R-L of 0.11, TR of 0.96, and the best NEK and GEK of 70.8\% and 72.3\%, showing stronger boundary preservation and general utility.

\section{Discussions}
\paragraph{Component ablation}
Table~\ref{tab_component_ablation} isolates each ADU component on Llama3.1-8B-Instruct. Removing the pathway loss raises forgetting metrics sharply, indicating that suppressing sensitive pathway mass drives forgetting. Removing the retain loss keeps forgetting but reduces MMLU and GSM8K by 6.52 and 6.72 points, showing that retain language modeling is essential for utility. Random heads weaken both forgetting and retention, suggesting that the local-global partition is non-interchangeable. Removing the sensitive anchor filter harms MMLU and GSM8K by penalizing benign high-APS anchors. Removing the RAS based preplan selection weakens forgetting and retention, confirming that ADU benefits from intervening at the transition point before sensitive anchors guide later generation. Full results including TOFU metrics are provided in Appendix D.2.

\paragraph{Path-specific causal validation.}
Figure~\ref{fig_temporal_rhythm}(b) visualizes how RAS peaks mark preplan transitions and APS peaks identify persistently attended anchors, localizing candidate pathways without proving they control sensitive retrieval. We therefore intervene directly on the pre-output-projection contributions of the identified preplan--anchor edges. Removing them from Base lowers WMDP Avg. from 58.62 to 36.37, whereas removing a cardinality-matched random edge set yields 57.09, showing retrieval depends specifically on the selected pathway rather than an arbitrary same-sized perturbation. Conversely, replacing ADU's selected contributions with their Base counterparts restores WMDP Avg. from 27.65 to 47.23, while matched-random replacement reaches only 28.82. The selected interventions change MMLU by only -0.58 and +0.53 points, respectively. These complementary results link ADU's training target to its behavioral effect: the selected contributions support a substantial portion of Base retrieval, and restoring their original computation recovers much of the access suppressed by ADU. Appendix~E provides complete bidirectional replacement analysis.

\begin{table}[t]
\centering
\resizebox{\columnwidth}{!}{
\begin{tabular}{lcc}
\toprule
\textbf{Condition} & \textbf{WMDP Avg.$\downarrow$}
& \textbf{MMLU$\uparrow$} \\
\midrule
Base                                           & 58.62 & 68.16 \\
Base \(-\) Selected \(\widetilde C_e\)         & \textbf{36.37} & 67.58 \\
Base \(-\) Matched random \(\widetilde C_e\)   & 57.09 & 67.52 \\
\midrule
ADU                                            & 27.65 & 62.84 \\
ADU \(\leftarrow\) Base selected               & {47.23} & 63.37 \\
ADU \(\leftarrow\) Base matched random         & 28.82 & 61.64 \\
\bottomrule
\end{tabular}}
\caption{Path-specific contribution interventions on
Llama3.1-8B-Instruct. ``\(\leftarrow\)'' replaces selected contributions
in the running model with their source-model counterparts.}
\label{tab:edge_causal_main}
\end{table}

\begin{figure}[t]
\centering
\includegraphics[width=0.9\linewidth]{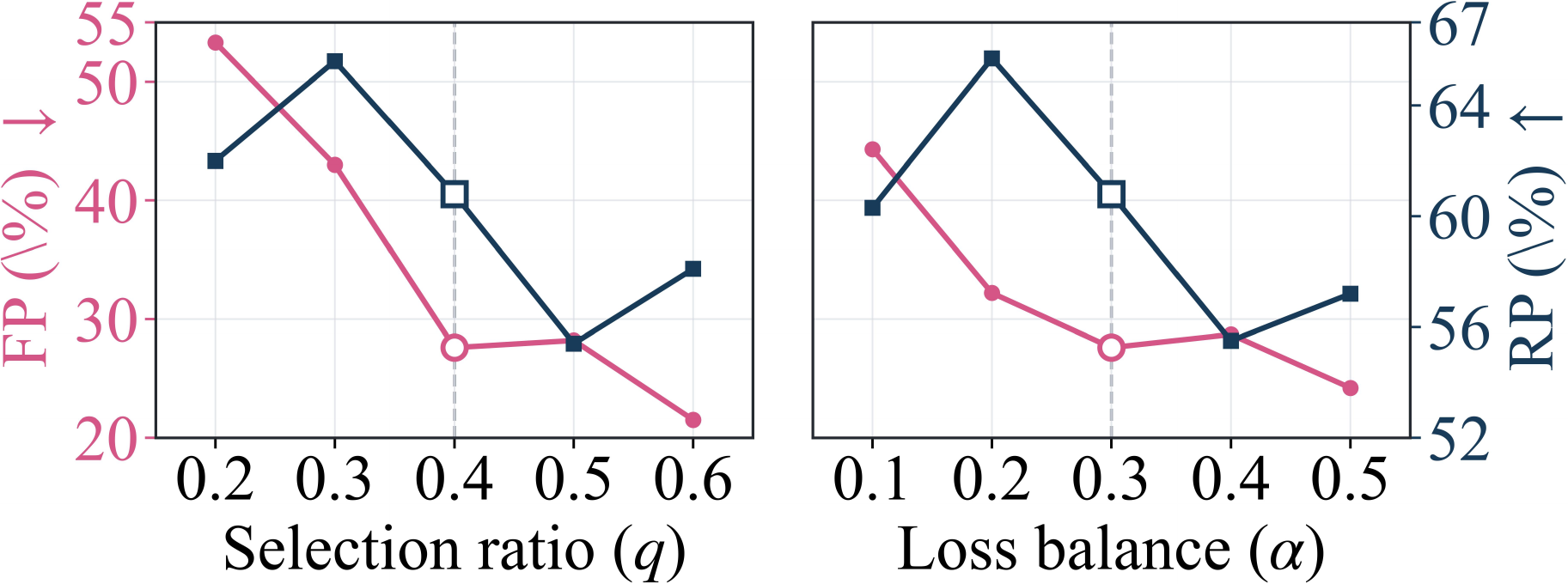}
\caption{Parameter sensitivity analysis on WMDP and retention tasks with Llama3.1-8B-Instruct.}
\label{fig_hyperparam_sensitivity}
\end{figure}

\paragraph{Hyperparameter Sensitivity Analysis}
We analyze the selection ratio \(q\), and the loss balance \(\alpha\) on Llama3.1-8B-Instruct. Figure~\ref{fig_hyperparam_sensitivity} reports the forgetting-retention trade-off when varying one hyperparameter while fixing the others to their default values. Small \(q\) misses sensitive pathways and leaves higher WMDP accuracy, whereas large \(q\) includes benign anchors and harms retention. The default \(q=0.4\) achieves FP 27.65 and RP 60.83, which forms a stable trade-off knee while avoiding the retention degradation observed at larger \(q\) values. A small \(\alpha\) underweights the pathway objective and generally weakens forgetting, whereas large \(\alpha\) provides limited forgetting gains and lowers retention performance. The default \(\alpha=0.3\) gives the best tested balance. Full numerical results and seed stability are reported in Appendix D.3 and Appendix D.4.

\begin{figure}[t]
\centering
\includegraphics[width=\linewidth]{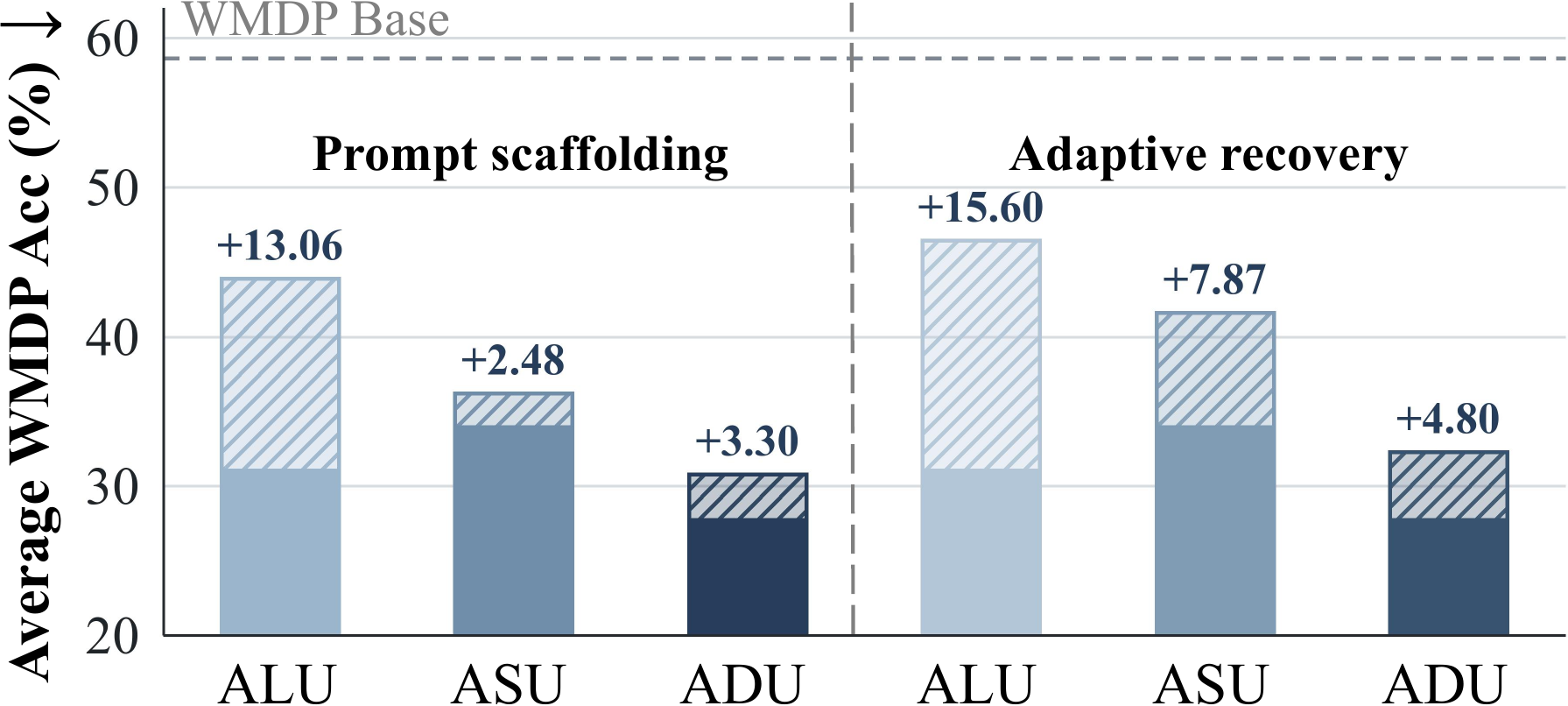}  
\caption{Groupwise worst-case WMDP accuracy under different attacks on Llama3.1-8B-Instruct.}
\label{fig_attack_robustness}
\end{figure}

\paragraph{Robustness Analysis.}
We group six attacks into prompt scaffolding (few-shot, masking, and
role-play CoT) and adaptive recovery (anchor shift, multi-turn probing,
and repeated sampling). They test whether altered reasoning contexts or
retrieval strategies re-elicit forgotten knowledge. As shown in
Fig.~\ref{fig_attack_robustness}, although ASU has a smaller increase
under prompt scaffolding, ADU still achieves the lowest attacked
accuracy. Under adaptive recovery, ADU has both the smallest increase
and lowest final accuracy, remaining below ALU and ASU. Thus, pathway
decoupling limits knowledge recovery beyond the original prompt. More
results and details are in Appendix~F.1.

\section{Conclusion}
Our work formulates LLM unlearning as contextual pathway decoupling:
sensitive knowledge is retrieved through context-dependent internal
routes and should not be reduced to an entire sequence, a static token,
or a prompt-level refusal. Based on this view, we introduce ADU, which
identifies a preplan--anchor rhythm from the temporal specialization of
local and global attention heads, fixes candidate retrieval paths under
the original model, and trains attention-projection adapters to suppress
them while preserving retain-set language modeling and local-attention
structure. The resulting framework provides a persistent parameter-level
mechanism that targets sensitive retrieval in context, while limiting
excessive forgetting, utility degradation, and recovery under altered
prompts. Experiments on WMDP, TOFU, and MUSE-Books demonstrate strong
forgetting--retention trade-offs, and bidirectional edge-contribution
interventions establish that the selected paths causally mediate a
substantial portion of sensitive retrieval and the learned forgetting
effect. Future work will extend pathway identification beyond attention,
improve automatic anchor construction, and develop stronger guarantees
against residual knowledge recovery.

\bibliography{aaai2027}

@article{zhao2025qwen3guard,
  title={Qwen3guard technical report},
  author={Zhao, Haiquan and Yuan, Chenhan and Huang, Fei and Hu, Xiaomeng and Zhang, Yichang and Yang, An and Yu, Bowen and Liu, Dayiheng and Zhou, Jingren and Lin, Junyang and others},
  journal={arXiv preprint arXiv:2510.14276},
  year={2025}
}

@article{sanyal2025agents,
  title={Agents are all you need for LLM unlearning},
  author={Sanyal, Debdeep and Mandal, Murari},
  journal={arXiv preprint arXiv:2502.00406},
  year={2025}
}

@InProceedings{Gong_2026_CVPR,
  author    = {Gong, Qinghui and Yang, Xue and Chen, Xunlei and Lai, Jinshan and Meng, Hua and Tang, Xiaohu},
  title     = {FedOrtho: Efficient Federated Unlearning Via Orthogonal Convolution and Adaptive Soft Pruning},
  booktitle = {Proceedings of the IEEE/CVF Conference on Computer Vision and Pattern Recognition (CVPR) Findings},
  month     = {June},
  year      = {2026},
  pages     = {8009--8018}
}

@inproceedings{shi2025muse,
  title={Muse: Machine unlearning six-way evaluation for language models},
  author={Shi, Weijia and Lee, Jaechan and Huang, Yangsibo and Malladi, Sadhika and Zhao, Jieyu and Holtzman, Ari and Liu, Daogao and Zettlemoyer, Luke and Smith, Noah and Zhang, Chiyuan},
  booktitle={International Conference on Learning Representations},
  volume={2025},
  pages={27797--27818},
  year={2025}
}

@inproceedings{lee2026direct,
  title={Direct Token Optimization: A Self-Contained Approach to Large Language Model Unlearning},
  author={Lee, Hong Kyu and Liu, Ruixuan and Xiong, Li},
  booktitle={Findings of the Association for Computational Linguistics: ACL 2026},
  pages={42083--42100},
  year={2026}
}

@article{kim2026unlearning,
  title={Unlearning-aware minimization},
  author={Kim, Hoki and Kim, Keonwoo and Chae, Sungwon and Yoon, Sangwon},
  journal={Advances in Neural Information Processing Systems},
  volume={38},
  pages={93806--93829},
  year={2026}
}

@inproceedings{pu2026decoding,
  title={Decoding-Unlearning: Fact Forgetting via Entropy-Guided Inference},
  author={Pu, Jingwen and Shi, Mingjun and Ren, Xinrui and Wang, Yizhe and Zhang, Xinyu and Wang, Zhaokun and She, Kun},
  booktitle={Proceedings of the 64th Annual Meeting of the Association for Computational Linguistics (Volume 1: Long Papers)},
  pages={39834--39860},
  year={2026}
}

@inproceedings{wang2026cap,
  title={CAP: Controllable Alignment Prompting for Unlearning in {LLM}s},
  author={Wang, Zhaokun and Guo, Jinyu and Pu, Jingwen and Pu, Hongli and Yang, Meng and Chen, Xunlei and Ou, Jie and Li, Wenyi and Luo, Guangchun and Tian, Wenhong},
  booktitle={Proceedings of the 64th Annual Meeting of the Association for Computational Linguistics (Volume 1: Long Papers)},
  year={2026}
}

@article{geng2025comprehensive,
  title={A comprehensive survey of machine unlearning techniques for large language models},
  author={Geng, Jiahui and Li, Qing and Woisetschlaeger, Herbert and Chen, Zongxiong and Cai, Fengyu and Wang, Yuxia and Nakov, Preslav and Jacobsen, Hans-Arno and Karray, Fakhri},
  journal={arXiv preprint arXiv:2503.01854},
  year={2025}
}

@inproceedings{ranjan2026razor,
  title={Razor: Ratio-aware layer editing for targeted unlearning in vision transformers and diffusion models},
  author={Ranjan, Ravi and Grover, Utkarsh and Lin, Xiaomin and Polyzou, Agoritsa},
  booktitle={Proceedings of the IEEE/CVF Conference on Computer Vision and Pattern Recognition},
  pages={7998--8008},
  year={2026}
}

@inproceedings{lin2025critical,
  title={Critical Tokens Matter: Token-Level Contrastive Estimation Enhances LLM’s Reasoning Capability},
  author={Lin, Zicheng and Liang, Tian and Xu, Jiahao and Liu, Qiuzhi and Wang, Xing and Luo, Ruilin and Shi, Chufan and Li, Siheng and Yang, Yujiu and Tu, Zhaopeng},
  booktitle={International Conference on Machine Learning},
  pages={37906--37918},
  year={2025},
  organization={PMLR}
}

@inproceedings{pawelczyk2024context,
  title={In-Context Unlearning: Language Models as Few-Shot Unlearners},
  author={Pawelczyk, Martin and Neel, Seth and Lakkaraju, Himabindu},
  booktitle={International Conference on Machine Learning},
  pages={40034--40050},
  year={2024},
  organization={PMLR}
}

@article{sun2025generative,
  title={Generative adversarial networks unlearning},
  author={Sun, Hui and Zhu, Tianqing and Chang, Wenhan and Zhou, Wanlei},
  journal={IEEE Transactions on Dependable and Secure Computing},
  year={2025},
  publisher={IEEE}
}

@inproceedings{yuan2025towards,
  title={Towards robust knowledge unlearning: An adversarial framework for assessing and improving unlearning robustness in large language models},
  author={Yuan, Hongbang and Jin, Zhuoran and Cao, Pengfei and Chen, Yubo and Liu, Kang and Zhao, Jun},
  booktitle={Proceedings of the AAAI Conference on Artificial Intelligence},
  volume={39},
  number={24},
  pages={25769--25777},
  year={2025}
}

@inproceedings{bhaila2025soft,
  title={Soft prompting for unlearning in large language models},
  author={Bhaila, Karuna and Van, Minh-Hao and Wu, Xintao},
  booktitle={Proceedings of the 2025 Conference of the Nations of the Americas Chapter of the Association for Computational Linguistics: Human Language Technologies (Volume 1: Long Papers)},
  pages={4046--4056},
  year={2025}
}

@inproceedings{cha2024learning,
  title={Learning to unlearn: Instance-wise unlearning for pre-trained classifiers},
  author={Cha, Sungmin and Cho, Sungjun and Hwang, Dasol and Lee, Honglak and Moon, Taesup and Lee, Moontae},
  booktitle={Proceedings of the AAAI conference on artificial intelligence},
  volume={38},
  number={10},
  pages={11186--11194},
  year={2024}
}

@article{hu2025exact,
  title={Exact and efficient unlearning for large language model-based recommendation},
  author={Hu, Zhiyu and Zhang, Yang and Xiao, Minghao and Wang, Wenjie and Feng, Fuli and He, Xiangnan},
  journal={IEEE Transactions on Knowledge and Data Engineering},
  year={2025},
  publisher={IEEE}
}

@inproceedings{zhao2025unlearning,
  title={Unlearning backdoor attacks for llms with weak-to-strong knowledge distillation},
  author={Zhao, Shuai and Wu, Xiaobao and Nguyen, Cong-Duy T and Jia, Yanhao and Jia, Meihuizi and Yichao, Feng and Tuan, Luu Anh},
  booktitle={Findings of the Association for Computational Linguistics: ACL 2025},
  pages={4937--4952},
  year={2025}
}

@inproceedings{jiang2025backdoor,
  title={Backdoor Token Unlearning: Exposing and Defending Backdoors in Pretrained Language Models},
  author={Jiang, Peihai and Lyu, Xixiang and Li, Yige and Ma, Jing},
  booktitle={Proceedings of the AAAI Conference on Artificial Intelligence},
  volume={39},
  number={23},
  pages={24285--24293},
  year={2025}
}

@inproceedings{chen2026alter,
  title={ALTER: Asymmetric loRA for token-entropy-guided unlearning of LLMs},
  author={Chen, Xunlei and Guo, Jinyu and Li, Yuang and Wang, Zhaokun and Gong, Yi and Zou, Jie and Wei, Jiwei and Tian, Wenhong},
  booktitle={Proceedings of the AAAI Conference on Artificial Intelligence},
  volume={40},
  number={42},
  pages={35366--35374},
  year={2026}
}

@article{cobbe2021training,
  title={Training verifiers to solve math word problems},
  author={Cobbe, Karl and Kosaraju, Vineet and Bavarian, Mohammad and Chen, Mark and Jun, Heewoo and Kaiser, Lukasz and Plappert, Matthias and Tworek, Jerry and Hilton, Jacob and Nakano, Reiichiro and others},
  journal={arXiv preprint arXiv:2110.14168},
  year={2021}
}

@inproceedings{tran-etal-2025-tokens,
    title = "Tokens for Learning, Tokens for Unlearning: Mitigating Membership Inference Attacks in Large Language Models via Dual-Purpose Training",
    author = "Tran, Toan  and
      Liu, Ruixuan  and
      Xiong, Li",
    editor = "Che, Wanxiang  and
      Nabende, Joyce  and
      Shutova, Ekaterina  and
      Pilehvar, Mohammad Taher",
    booktitle = "Findings of the Association for Computational Linguistics: ACL 2025",
    month = jul,
    year = "2025",
    address = "Vienna, Austria",
    publisher = "Association for Computational Linguistics",
    url = "https://aclanthology.org/2025.findings-acl.1174/",
    doi = "10.18653/v1/2025.findings-acl.1174",
    pages = "22872--22888",
    ISBN = "979-8-89176-256-5"
}

@article{tan2025wisdom,
  title={Wisdom is Knowing What not to Say: Hallucination-Free LLMs Unlearning via Attention Shifting},
  author={Tan, Chenchen and Qu, Youyang and Li, Xinghao and Zhang, Hui and Cui, Shujie and Chen, Cunjian and Gao, Longxiang},
  journal={NeurIPS 2025},
  year={2025}
}

@inproceedings{li2025forget,
  title={Forget the Token and Pixel: Rethinking Gradient Ascent for Concept Unlearning in Multimodal Generative Models},
  author={Li, Jiaqi and Zhang, Chuanyi and Du, Miaozeng and Zhang, Hui and Chen, Yongrui and Wei, Qianshan and Fang, Junfeng and Wang, Ruipeng and Bi, Sheng and Qi, Guilin},
  booktitle={Findings of the Association for Computational Linguistics: ACL 2025},
  pages={12179--12200},
  year={2025}
}

@article{yu2025unierase,
  title={UniErase: Unlearning Token as a Universal Erasure Primitive for Language Models},
  author={Yu, Miao and others},
  journal={arXiv preprint arXiv:2505.15674},
  year={2025}
}

@inproceedings{shah2025unlearning,
  title={The unlearning mirage: A dynamic framework for evaluating LLM unlearning},
  author={Shah, Raj Sanjay and Huang, Jing and Murugesan, Keerthiram and Baracaldo, Nathalie and Yang, Diyi},
  booktitle={Second Conference on Language Modeling},
  year={2025}
}

@inproceedings{thudi2022unrolling,
  title={Unrolling sgd: Understanding factors influencing machine unlearning},
  author={Thudi, Anvith and Deza, Gabriel and Chandrasekaran, Varun and Papernot, Nicolas},
  booktitle={EuroS\&P 2022},
  pages={303--319},
  year={2022},
  organization={IEEE}
}

@inproceedings{liu2025disentangling,
  title={Disentangling biased knowledge from reasoning in large language models via machine unlearning},
  author={Liu, Zheyuan and Maharjan, Suraj and Wu, Fanyou and Parikh, Rahil and Bayar, Belhassen and Sengamedu, Srinivasan H and Jiang, Meng},
  booktitle={Proceedings of the 63rd Annual Meeting of the Association for Computational Linguistics (Volume 1: Long Papers)},
  pages={6105--6123},
  year={2025}
}

@inproceedings{wang2025selective,
  title={Selective forgetting: Advancing machine unlearning techniques and evaluation in language models},
  author={Wang, Lingzhi and Zeng, Xingshan and Guo, Jinsong and Wong, Kam-Fai and Gottlob, Georg},
  booktitle={Proceedings of the AAAI Conference on Artificial Intelligence},
  volume={39},
  number={1},
  pages={843--851},
  year={2025}
}

@inproceedings{wang2025llm,
  title={LLM Unlearning via Loss Adjustment with Only Forget Data},
  author={Wang, Yaxuan and Wei, Jiaheng and others},
  booktitle={The Thirteenth International Conference on Learning Representations},
  year={2025},
}

@inproceedings{zhang2024negative,
    title={Negative Preference Optimization: From Catastrophic Collapse to Effective Unlearning},
    author={Ruiqi Zhang and Licong Lin and Yu Bai and Song Mei},
    booktitle={First Conference on Language Modeling},
    year={2024},
    url={https://openreview.net/forum?id=MXLBXjQkmb}
}

@article{Shi2024Safety,
  publtype={informal},
  author={Dan Shi and others},
  title={Large Language Model Safety: A Holistic Survey},
  year={2024},
  cdate={1704067200000},
  journal={CoRR},
  volume={abs/2412.17686},
  url={https://doi.org/10.48550/arXiv.2412.17686}
}

@article{maini2024tofu,
  title={Tofu: A task of fictitious unlearning for llms},
  author={Maini, Pratyush and Feng, Zhili and Schwarzschild, Avi and Lipton, Zachary C and Kolter, J Zico},
  journal={arXiv preprint arXiv:2401.06121},
  year={2024}
}

@article{Grynbaum2023Times,
  title={The Times sues OpenAI and Microsoft over AI use of copyrighted work},
  author={Grynbaum, Michael M and others},
  journal={The New York Times},
  volume={27},
  number={1},
  year={2023}
}

@article{Lee2020BioBERT,
  title={BioBERT: a pre-trained biomedical language representation model for biomedical text mining},
  author={Lee, Jinhyuk and others},
  journal={Bioinformatics},
  volume={36},
  number={4},
  pages={1234--1240},
  year={2020},
  publisher={Oxford University Press}
}

@article{zhao2024makes,
  title={What makes unlearning hard and what to do about it},
  author={Zhao, Kairan and Kurmanji, Meghdad and B{\u{a}}rbulescu, George-Octavian and Triantafillou, Eleni and Triantafillou, Peter},
  journal={Advances in Neural Information Processing Systems},
  volume={37},
  pages={12293--12333},
  year={2024}
}

@article{nguyen2025survey,
  title={A survey of machine unlearning},
  author={Nguyen, Thanh Tam and Huynh, Thanh Trung and Ren, Zhao and Nguyen, Phi Le and Liew, Alan Wee-Chung and Yin, Hongzhi and Nguyen, Quoc Viet Hung},
  journal={ACM Transactions on Intelligent Systems and Technology},
  volume={16},
  number={5},
  pages={1--46},
  year={2025},
  publisher={ACM New York, NY}
}

@inproceedings{yang2025cliperase,
  title={Cliperase: Efficient unlearning of visual-textual associations in clip},
  author={Yang, Tianyu and Dai, Lisen and Wang, Xiangqi and Cheng, Minhao and Tian, Yapeng and Zhang, Xiangliang},
  booktitle={Proceedings of the 63rd Annual Meeting of the Association for Computational Linguistics (Volume 1: Long Papers)},
  pages={30438--30452},
  year={2025}
}

@inproceedings{li2025wmdp,
  title={The WMDP Benchmark: Measuring and Reducing Malicious Use with Unlearning},
  author={Li, Nathaniel and Pan, Alexander and Gopal, Anjali and Yue, Summer and Berrios, Daniel and Gatti, Alice and Li, Justin D and Dombrowski, Ann-Kathrin and Goel, Shashwat and Mukobi, Gabriel and others},
  booktitle={International Conference on Machine Learning},
  pages={28525--28550},
  year={2025},
  organization={PMLR}
}

@inproceedings{zhuang2025seuf,
  title={SEUF: Is Unlearning One Expert Enough for Mixture-of-Experts LLMs?},
  author={Zhuang, Haomin and Zhang, Yihua and Guo, Kehan and Jia, Jinghan and Liu, Gaowen and Liu, Sijia and Zhang, Xiangliang},
  booktitle={Proceedings of the 63rd Annual Meeting of the Association for Computational Linguistics (Volume 1: Long Papers)},
  pages={8664--8678},
  year={2025}
}

@inproceedings{jin2025disentangling,
  title={Disentangling memory and reasoning ability in large language models},
  author={Jin, Mingyu and Luo, Weidi and Cheng, Sitao and Wang, Xinyi and Hua, Wenyue and Tang, Ruixiang and Wang, William Yang and Zhang, Yongfeng},
  booktitle={Proceedings of the 63rd Annual Meeting of the Association for Computational Linguistics (Volume 1: Long Papers)},
  pages={1681--1701},
  year={2025}
}

@article{shi2024muse,
  title={MUSE: Machine Unlearning Six-Way Evaluation for Language Models},
  author={Shi, Weijia and Malladi, Sadhika and Zhao, Jieyu and Holtzman, Ari and Liu, Daogao and Zettlemoyer, Luke and Smith, Noah A. and Zhang, Chiyuan},
  journal={arXiv preprint arXiv:2407.06460},
  year={2024}
}

@inproceedings{hendrycks2021measuring,
    title={Measuring Massive Multitask Language Understanding},
    author={Dan Hendrycks and Collin Burns and Steven Basart and Andy Zou and Mantas Mazeika and Dawn Song and Jacob Steinhardt},
    booktitle={International Conference on Learning Representations},
    year={2021},
    url={https://openreview.net/forum?id=d7KBjmI3GmQ}
}


\end{document}